\documentclass[runningheads]{llncs}
\usepackage{ dsfont }
\usepackage{booktabs,colortbl}
\usepackage{subcaption}
\usepackage[pdftex]{graphicx}
\usepackage{amsmath,bm}
\usepackage{amssymb}
\usepackage{cleveref}
\usepackage{wrapfig}
\usepackage{enumerate}
\usepackage[shortlabels]{enumitem}
\usepackage{xcolor}
\usepackage{comment}
\usepackage[misc]{ifsym}
\usepackage[hyphens]{url}
\begin{document}

% ECML 
% The maximum length of papers is 14 pages for the technical content excluding references in this format. 

\title{How To Overcome Confirmation Bias in Semi-Supervised Image Classification By Active Learning}

\titlerunning{Overcome Conf. Bias in SSL by AL}
% If the paper title is too long for the running head, you can set
% an abbreviated paper title here
%
\author{Sandra Gilhuber$^*$\inst{1,2}(\Letter) \and
Rasmus Hvingelby$^*$\inst{3} \and
Mang Ling Ada Fok\inst{3} \and
Thomas Seidl\inst{1,2,3}}
% %
\authorrunning{S. Gilhuber \& R. Hvingelby et al.}
% First names are abbreviated in the running head.
% If there are more than two authors, 'et al.' is used.
%
\institute{LMU Munich, Germany 
\email{$\{$gilhuber,seidl$\}$@dbs.ifi.lmu.de}
\and 
Munich Center for Machine Learning (MCML), Germany
\and
Fraunhofer IIS, Germany 
\email{rasmus.hvingelby@iis.fraunhofer.de}}

\toctitle{How To Overcome Confirmation Bias in Semi-Supervised Image Classification By Active Learning}
\tocauthor{Sandra~Gilhuber \& Rasmus~Hvingelby}

\maketitle              % typeset the header of the contribution
\def\thefootnote{*}\footnotetext{Equal contribution}\def\thefootnote{\arabic{footnote}}
\setcounter{footnote}{0}

\begin{abstract}
     Do we need active learning? The rise of strong deep semi-supervised methods raises doubt about the usability of active learning in limited labeled data settings. This is caused by results showing that combining semi-supervised learning (SSL) methods with a random selection for labeling can outperform existing active learning (AL) techniques. However, these results are obtained from experiments on well-established benchmark datasets that can overestimate the external validity. However, the literature lacks sufficient research on the performance of active semi-supervised learning methods in realistic data scenarios, leaving a notable gap in our understanding. Therefore we present three data challenges common in real-world applications: between-class imbalance, within-class imbalance, and between-class similarity. These challenges can hurt SSL performance due to confirmation bias. We conduct experiments with SSL and AL on simulated data challenges and find that random sampling does not mitigate confirmation bias and, in some cases, leads to worse performance than supervised learning. In contrast, we demonstrate that AL can overcome confirmation bias in SSL in these realistic settings. Our results provide insights into the potential of combining active and semi-supervised learning in the presence of common real-world challenges, which is a promising direction for robust methods when learning with limited labeled data in real-world applications.

\end{abstract}

\section{Introduction}
The success of supervised deep learning models largely depends on the availability of sufficient, qualitative labeled data. Since manual annotation is time-consuming and costly, various research directions focus on machine learning with limited labeled data. While Active Learning (AL)~\cite{beck2021effective,settles2009active} aims to label only the most informative and valuable data intelligently, semi-supervised learning (SSL)~\cite{berthelot2019mixmatch,chapelle2009semi,sohn2020fixmatch} aims to exploit the information in the unlabeled pool without asking for new labels. 
Given the complementary nature of SSL and AL, it is intuitive to explore their integration within a unified framework to maximize the utilization of the available data. However, the effectiveness of AL has been questioned recently~\cite{mittal2019parting,chan2020marginal,bengar2021reducing,lowell2019practical}. Some works show that other learning paradigms capable of exploiting the unlabeled data do not experience added value from biased and intelligent data selection through AL~\cite{chan2020marginal}. 

However, these findings are mainly based on experiments on well-established, clean benchmark datasets. But, an excessive emphasis on benchmark performance can result in diminishing returns where increasingly large efforts lead to ever-decreasing performance gains on the actual task \cite{liao2021are,varoquaux2022machine}.
As a result, an exclusive evaluation of such benchmarks can raise concerns about the transferability of these results to challenges in real-world applications. Therefore, we review the literature on AL to understand which datasets are commonly used for evaluation and to what extent AL has been combined with SSL.

Toward a better understanding, we first categorize existing AL methods into four groups, namely uncertainty sampling, representativeness sampling, coverage-based sampling, and balanced sampling. Second, we introduce the following three real-world challenges: 
(1) \emph{Between-class imbalance} (BCI), where the distribution over class instances is non-uniform, (2) \emph{within-class imbalance} (WCI), where the intra-class distribution is non-uniform, and (3) \emph{between-class similarity} (BCS), where the class boundaries are ambiguous. In our experiments, we demonstrate that each of these real-world challenges introduces confirmation bias reinforcing biased or misleading concepts toward SSL. Moreover, randomly increasing the labeled pool may not effectively address the posed challenges. In fact, the results stagnate early or are even worse than plain supervised learning. In contrast, we evaluate simple AL heuristics on the introduced challenges and show that active data selection leads to much better generalization performance in these cases. 
This provides empirical evidence of the benefits of incorporating AL techniques to mitigate the impact of real-world challenges in SSL.

Our main contributions are: 
\begin{itemize}
    \item We provide a thorough literature review on the real-world validity of current evaluation protocols for active and semi-supervised learning. We find that the combination is especially understudied in real-world datasets. 
    \item We explore well-established SSL methods in three real-world challenges and find that confirmation bias in SSL is a problem in all studied challenges and leads to degraded performance. 
    \item We show that, in contrast to random selection, \emph{actively} increasing the labeled pool can mitigate these problems.
\end{itemize}

\section{Related Work}
\label{sec:literature}
The advantages of AL have been questioned due to the strong performance of methods exploiting knowledge available in unlabeled data~\cite{mittal2019parting,chan2020marginal,bengar2021reducing}. 

Given AL aims to increase model performance while decreasing annotation efforts, it is important not to focus on AL in isolation when other training techniques can lead to improvements in model performance. This makes the evaluation of AL challenging~\cite{luth2023realistic} as there are many ways to configure AL, and it can be hard to know upfront what works in a real-world scenario.

Our focus is specifically on three realistic data scenarios that can lead SSL to underperform due to confirmation bias.

\subsection{Real World Considerations in Machine Learning}
The evaluation of the algorithmic progress on a task can be separated into \emph{internal} validity and \emph{external} validity~\cite{liao2021are}. When benchmark results are internally valid, the improvements caused by an algorithm are valid within the same dataset. However, the overuse of the same test sets in benchmarks can lead to adaptive overfitting where the models and hyperparameters yielding strong performance are reused, and the improvements are not necessarily caused by algorithmic improvements.
On the other hand, external validity refers to whether improvements also translate to other datasets for the same task. It has been observed that an excessive emphasis on benchmark performance can result in diminishing returns where increasingly large efforts lead to smaller and smaller performance gains on the actual task \cite{liao2021are,varoquaux2022machine}. To improve the validity of benchmark results, it is important that the datasets used for evaluation reflect the data challenges that occur in real-world scenarios.

Considering data challenges has been a well-studied field in machine learning. Lopez et al. \cite{lopez2013insight} investigate how data intrinsic characteristics in imbalanced datasets affect classification performance and specify six problems that occur in real-world data. Both \cite{stefanowski2016dealing} and \cite{wojciechowski2017difficulty} also focus on imbalanced data and discuss difficulty factors that deteriorate classification performance. \cite{stefanowski2016dealing} further demonstrates that these factors have a larger impact than the imbalance ratio or the size of the minority class. \cite{das2018handling} investigates data irregularities that can lead to a degradation in classification performance. 
However, to the best of our knowledge studying data challenges in limited labeled scenarios has not yet been well studied \cite{oliver2018realistic,wang2022usb,luth2023realistic}.

\subsection{Evaluation of AL in the Literature}
To get an understanding of the data commonly used for evaluation in limited labels scenarios, we performed a literature overview of the papers published in 13 top-venue conferences\footnote{ACL, AAAI, CVPR, ECCV, ECML PKDD, EMNLP, ICCV, ICDM, ICLR, ICML, IJCAI, KDD, and NeurIPS} within Artificial Intelligence, Machine Learning, Computer Vision, Natural Language Processing and Data Mining between 2018 and 2022. We selected papers for screening if "\textit{active learning}" occurs in the title and abstract, resulting in 392 papers. When screening, we included papers that empirically study the improvement of machine learning models for image classification when expanding the pool of labeled data, as is common in AL papers. Based on this inclusion criteria, we first screened the title and abstracts, and if we could not exclude a study only on the title and abstract, we did a full-text screening. Following this screening process, we identified 51 papers.

We find that 47 (94\%) of the studies experimented on at least one benchmark dataset, and 38 (75\%) of the studies experiments solely on benchmark datasets\footnote{We consider benchmark datasets as the well-established MNIST, CIFAR10/100, SVHN, FashionMNIST, STL-10, ImageNet (and Tiny-ImageNet), as well as Caltech-101 and Caltech-256.}. To understand how common it is to evaluate AL in more realistic data scenarios, we count how many papers consider the data challenges BCI, WCI, or BCS or experiments on non-benchmark datasets. We find that 23 (45\%) papers consider real-world data challenges or evaluate non-benchmark datasets. The most common data challenge is BCI which 15 (29\%) of the papers are considering.
As AL can be improved with other training techniques, we look at how many papers combine AL and SSL and find that this is done by 13 (25\%) of the papers. However, only 5 (10\%) evaluate the performances in realistic scenarios. An overview can be found in the Appendix.

\section{Learning with Limited Labeled Data}
\label{sec:limited}
Given an input space $\mathcal{X}$ and a label space $\mathbf{Y}$, we consider the limited labeled scenario where we assume a small labeled pool $\mathcal{X}^l \subset \mathcal{X}$ and a large unlabeled data pool $\mathcal{X}^u = \mathcal{X} \setminus \mathcal{X}^l$. We want to obtain a model $f(x;\theta) \rightarrow \mathbb{R}^C$ where parameters $\theta$ map a given input $x \in \mathcal{X}$ to a $C$-dimensional vector. Supervised learning trains a model on $\mathcal{X}^l$ while SSL utilizes both $\mathcal{X}^l$ and $\mathcal{X}^u$.

\subsection{Semi-Supervised Learning (SSL)}
Many approaches to leverage both labeled and unlabeled data have been suggested in the literature~\cite{chapelle2009semi,van2020survey}. More recently, the utilization of deep learning in SSL has shown impressive performance, and especially different variants of consistency regularization and pseudo-labeling have been studied \cite{wang2022usb}.

\textbf{Pseudo-labeling}~\cite{lee2013pseudo} uses the model's prediction on the instances in $\mathcal{X}^u$ to filter highly confident samples and include those with their respective pseudo-label in the next training iteration. 
Pseudo-labeling is a simple and powerful technique for utilizing $\mathcal{X}^u$. However, a model producing incorrect predictions reuses wrong information in training. This is known as confirmation bias \cite{arazo2020pseudo} and can greatly impact model performance.

\textbf{Consistency Regularization}~\cite{sohn2020fixmatch,berthelot2019mixmatch} exploits $\mathcal{X}^u$ by encouraging invariant predictions when the input is perturbated, thereby making the model robust to different perturbed versions of unlabeled data. Perturbations of the data can be obtained by introducing random noise to the input data or utilizing data augmentations \cite{sohn2020fixmatch}. Some methods rely heavily on data augmentations which assume that label-preserving data augmentations are available when applying such methods in real-world use cases. Using consistency regularization in combination with pseudo-labeling helps improve the generalizability through the perturbed data, which can further enforce the confirmation bias if the model predictions are wrong.

\subsection{Active Learning (AL)}
\label{subsec:al}
AL alternates between querying instances for annotation, and re-training the model $f(x;\theta)$ on the increased labeled pool
until an annotation budget is exhausted or a certain performance is reached. The so-called acquisition function of an AL strategy determines which instances in $\mathcal{X}^u$ are most valuable and should be labeled to maximize the labeling efficiency.
We use the following taxonomy to distinguish between active acquisition types.

\begin{wrapfigure}{r}{0.4\textwidth}
  % \vspace{-15pt}
  \begin{center}
    \includegraphics[width=0.4\textwidth]{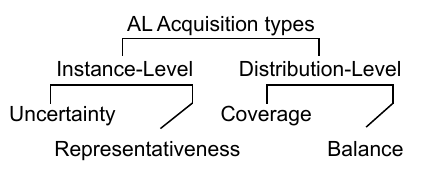}
  \end{center}
  \caption{AL Acquisition Types}
  % \vspace{-15pt}
\end{wrapfigure}

\textbf{Instance-Level Acquisition} 
Each unlabeled instance $x \in \mathcal{X}^u$ is assigned a scoring individually, independent of already selected instances, and enables a final ranking of all unlabeled instances.  

\begin{figure}[ht]
    \centering
    \begin{subfigure}[b]{0.4\textwidth}
        \includegraphics[width=\textwidth]{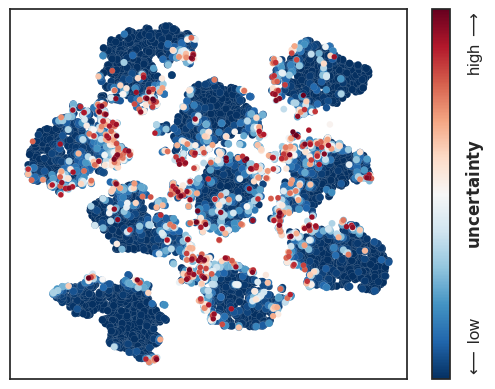}
        \caption{\textbf{Uncertainty Sampling}}
        \label{fig:unc}
    \end{subfigure}
    \hfill
    \begin{subfigure}[b]{0.4\textwidth}
        \includegraphics[width=\textwidth]{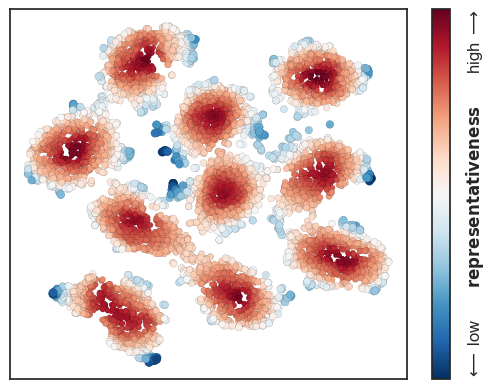}
        \caption{\textbf{Representative Sampling}}
        \label{fig:repr}
    \end{subfigure}
    \label{fig:informativeness}
    
    \begin{subfigure}[b]{0.4\textwidth}
        
        \includegraphics[width=\textwidth]{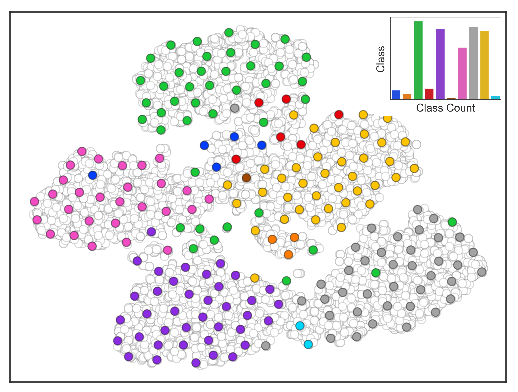}
        \caption{\textbf{Coverage-Based Sampling}}
        \label{fig:cover}
    \end{subfigure}
    \hfill
    \begin{subfigure}[b]{0.4\textwidth}
        \includegraphics[width=\textwidth]{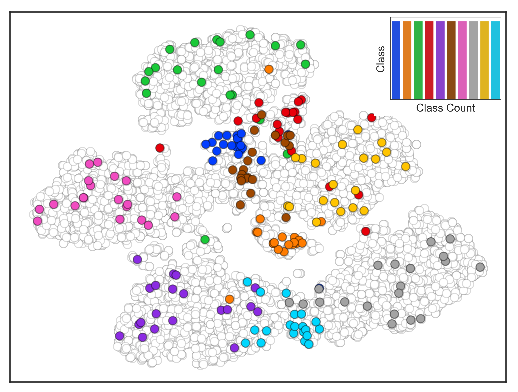}
        \caption{\textbf{Balanced Sampling}}
        \label{fig:balance}
    \end{subfigure}
    \caption{Exemplary illustration of different acquisition types.}
\end{figure}

\emph{Uncertainty sampling} aims to query instances carrying the most novel information for the current learner. Popular estimates are least-confidence, min margin, or max entropy selection~\cite{settles2009active}. These methods usually query near the class boundaries as illustrated in \Cref{fig:unc}. The 2D t-SNE visualization of MNIST shows a mapping of margin uncertainty, where red indicates high and blue indicates low uncertainty.
Other methods aim to measure model confidence~\cite{gal2017deep} and to distinguish between aleatoric and epistemic uncertainty~\cite{mukhoti2021deterministic}.

\emph{Representative sampling} assigns higher scores to instances representative of their class or a certain local region. The central idea is not to select instances to eliminate knowledge gaps in the current learning phase but to find instances that have the \emph{highest impact} on most other instances because, e.g., they are representative of a class or they are similar to many other instances. 
 
One way to define representativeness is to measure centrality, for instance, by exploiting a preceding partitioning and selecting the most central instance of each partitioning~\cite{prabhu2021active,zhdanov2019diverse}.  
Another estimate for representativeness is density, i.e., how many (similar) instances are in the near surrounding of a data point~\cite{donmez2007dual,wang2017active}. 
In \Cref{fig:repr}, the colors indicate the negative local outlier score~\cite{breunig2000lof} mapped onto the 2D representation of MNIST, which is here used as an indicator for representativeness. A representativeness selection strategy would favor instances in the denser red regions in the center of the clusters.

\textbf{Distribution-Level Acquisition} 
In contrast to instance-level, distribution-level acquisition refers to selection strategies that do not consider individual scores for each instance but strive to optimize the distribution of all selected instances. A clear ranking is usually not possible because the worthiness of the next best instance depends on which instance(s) is (have been) selected before.

\emph{Coverage-based sampling}, sometimes referred to as diversity sampling, aims to cover the given data space to avoid overlap of information best. The goal is to select as diverse instances as possible to maximize the richness of information in the labeled dataset. The most prominent method of this category is k-Center-Greedy which maximizes the distance in the feature space between the queried and the labeled instances ~\cite{sener2018active}. Coverage, or diversity, is a popular companion in hybrid approaches to assist batch-selection acquisitions~\cite{badge,kirsch2019batchbald,prabhu2021active}.  

\emph{Balanced sampling} aims to balance the number of samples per class and is especially suited for imbalanced datasets.
This subtype is often combined with other acquisition types, as it does not necessarily select the most valuable instances on its own~\cite{aggarwal2020active,bengar2022class,ertekin2007learning}. 
\Cref{fig:cover} depicts coverage sampling on an imbalanced version of MNIST where the data space is evenly covered. In contrast, \Cref{fig:balance} shows balanced sampling where the selected class counts are uniformly distributed. 

There is an abundance of hybrid methods combining two or more of the described concepts~\cite{badge,fu2021transferable,kirsch2019batchbald,prabhu2021active,xie2022towards}. However, in this work, we focus on highlighting the potential of AL in general and only consider disjoint baseline methods from each category. For an overview of deep AL methods, we refer to ~\cite{ren2021survey,wu2022deep,zhan2022comparative}.

\section{Three Real-World Data Challenges}
In the following, we introduce three realistic data challenges. We then present three datasets that demonstrate these challenges on the well-known MNIST task, which we later analyze in our experiments.

\begin{figure}[ht]
    \centering
    \hfill
    \begin{subfigure}[b]{0.3\textwidth}
        \includegraphics[width=\textwidth]{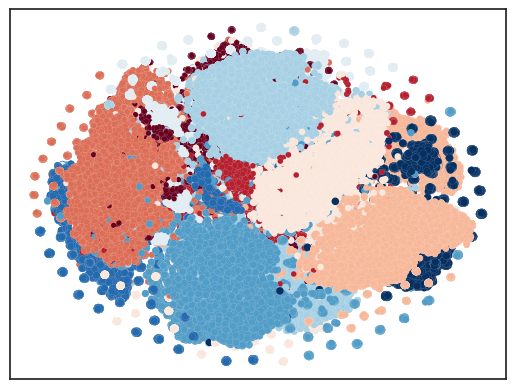}
        \caption{t-SNE BCI-MNIST.}
        \label{fig:bci_tsne}
    \end{subfigure}
    \hfill
    \begin{subfigure}[b]{0.3\textwidth}
        \includegraphics[width=\textwidth]{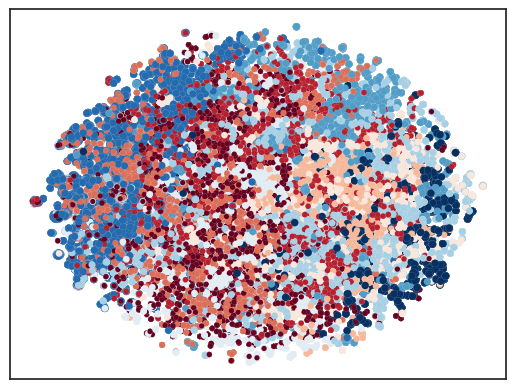}
        \caption{t-SNE BCS-MNIST.}
        \label{fig:bcs_tsne}
    \end{subfigure}
    \hfill  
    \begin{subfigure}[b]{0.3\textwidth}
        \includegraphics[width=\textwidth]{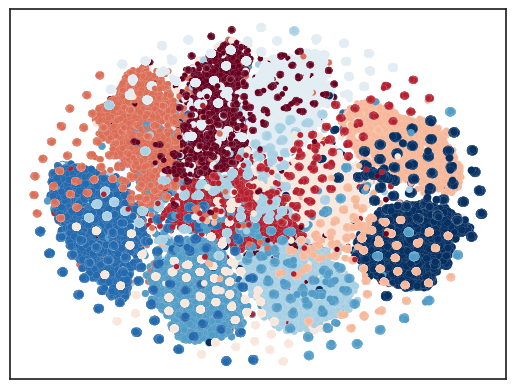}
        \caption{t-SNE WCI-MNIST.}
        \label{fig:wci_tsne}
    \end{subfigure}
    \begin{subfigure}[b]{0.3\textwidth}
        \includegraphics[width=\textwidth]{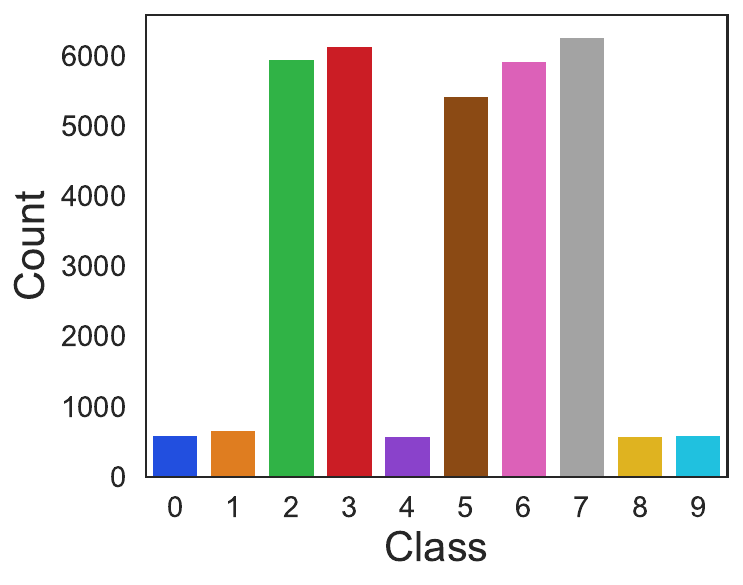}
        \caption{Class distribution BCI-MNIST.}
        \label{fig:bci_dist}
    \end{subfigure}
    \hfill 
    \begin{subfigure}[b]{0.32\textwidth}
        \includegraphics[width=\textwidth]{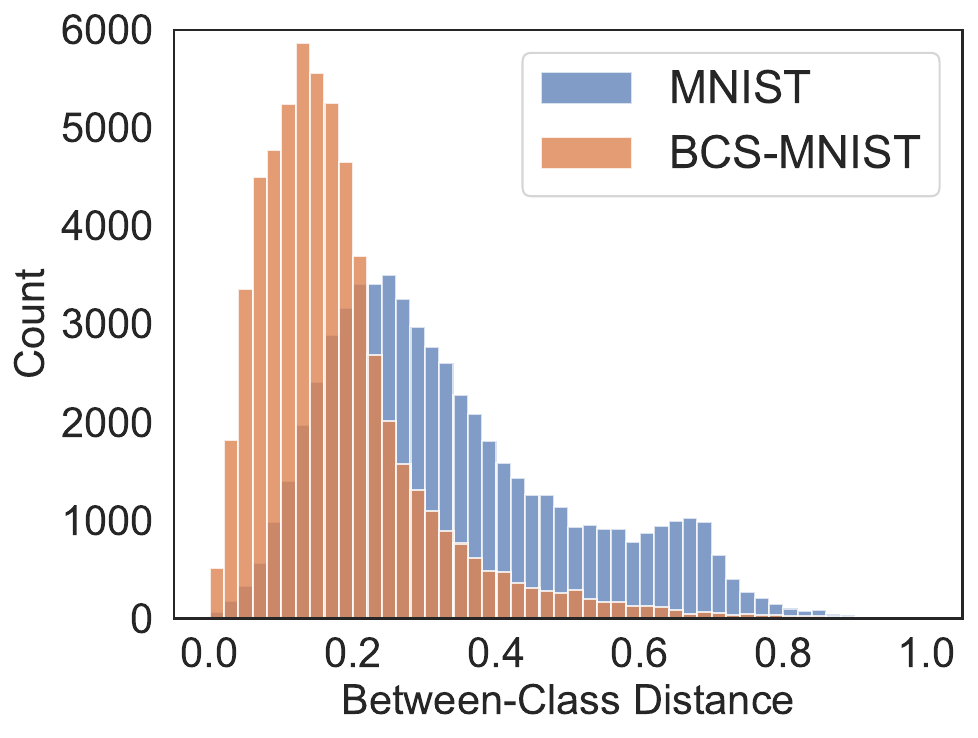}
        \caption{Between-class distances of BCS-MNIST vs. MNIST.}
        \label{fig:bcs_dist}
    \end{subfigure}
    \hfill
    \begin{subfigure}[b]{0.32\textwidth}
        \includegraphics[width=\textwidth]{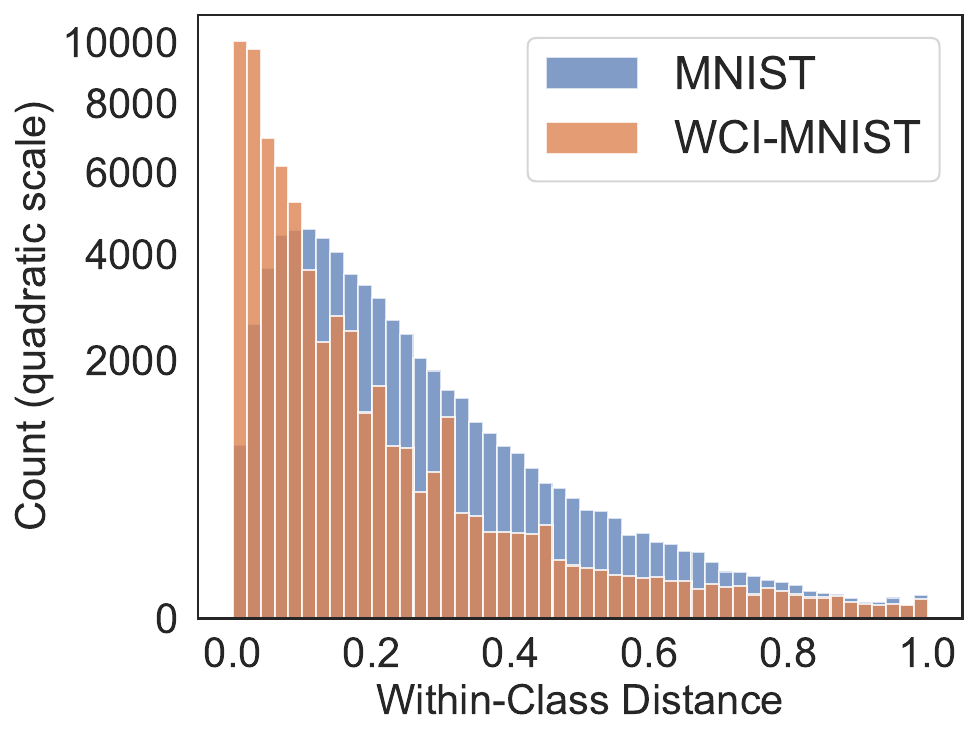}
        \caption{Within-class distances of WCI-MNIST vs. MNIST.}
        \label{fig:wci_dist}
    \end{subfigure}
    
    \caption{Three realistic challenges (BCI, BCS, WCI) demonstrated on MNIST.}
    \label{fig:ds_challenges}
\end{figure}

\subsection{Between-Class Imbalance (BCI)}
Among our challenges, Between-Class Imbalance (BCI) is the most considered in the literature and is a well-known challenge for supervised machine learning models. Imbalanced class distributions pose a problem for SSL methods where unlabeled data is often assumed to be distributed similarly to the labeled data and balanced class distributions. BCI can pose a problem for SSL when there is a mismatch between the labeled and unlabeled class distributions \cite{oliver2018realistic} or simply because some classes are generally underrepresented in both the unlabeled and labeled pool \cite{hyun2020class}.  
However, class distributions in real-world datasets often follow a long-tail distribution. 
While class imbalance has been studied for both AL and SSL separately, an open question remains regarding how to leverage AL techniques to address the negative effects of class imbalance in SSL.

\subsection{Between-Class Similarity (BCS)}
Another category of data challenges is Between-Class Similarity (BCS). In real-world datasets, the boundaries between classes can be hard to draw. Instances within the same class can differ widely, and conversely, instances from different classes can be very similar. High within-class diversity and similarity between classes happens naturally in many image classification tasks, e.g., diatom or plankton classification~\cite{venkataramanan2021tackling} or within histopathology~\cite{su2021semi}.

Datasets with BCS are a challenge for techniques that rely on unlabeled data for model training, 
since that contradicts the basic assumptions of SSL. 
For instance, according to \cite{chang2022viable}, 
Fixmatch exacerbates confusion when instances across classes are similar. The degree of BCS determines whether it is advantageous to sample from class boundaries while the classes can still be differentiated or to prioritize selecting representative instances without ambiguity in the class assignment. Consequently, this challenge presents an opportunity for AL to identify and label such samples.
This problem does not only occur on hard-to-solve tasks with high aleatoric uncertainty. 
Ambiguous label information can also occur due to the labeling procedure e.g. when data is labeled by multiple annotators which can introduce labeling variations~\cite{plank2022problem}, or when labels are acquired automatifcally~\cite{li2019learning,varoquaux2022machine}.
Label noise can have a large impact on SSL as the model is more prone to confirm learned mistakes leading to confirmation bias~\cite{li2020dividemix}.

Common usage of SSL methods for noisily labeled data is to simply remove noisy labels and continue training with conventional SSL~\cite{algan2021image}. Alternatively, some algorithms distinguish between cleanly labeled, noisily labeled, and unlabeled data enabling the usage of a massive amount of unlabeled and noisy data under the supervision of a few cleanly annotated data.
However, directly coupling the data selection actively to the training can be an easy and thus attractive solution to directly account for label noise or ambiguous class labels without post-processing wrong labels or complex algorithms and wasted labeling efforts.

\subsection{Within-Class Imbalance (WCI)}
Imbalance is not only a problem across classes but also within classes \cite{japkowicz2001concept-learning,huan2019resolving}. Although instances might belong to the same class, they can have a high variability due to, e.g., pose, lighting, viewpoint, etc. To obtain a model with the most discriminative capabilities, it must be exposed to the variation within the class. 

Within-class imbalance (WCI) occurs in many real-world problems. In medical imaging, subgroups such as race or gender exist within classes and are often imbalanced \cite{wang2019wgan}. Similarly, in microscopic classification, the images might have different viewpoints forming diverse~\cite{venkataramanan2021tackling} and imbalanced~\cite{lee2016plankton} subclusters. In automatic defect detection for manufacturing systems, the different types of defects are often all grouped into the same superordinate class and can be very diverse and imbalanced~\cite{huan2019resolving}.
It has also been shown that repetition of subclasses containing highly similar samples occurs in commonly used image classification benchmark datasets \cite{birodkar2019semantic}, leading to some subclasses that contain redundant semantic information being overrepresented.

WCI, similar to BCI, leads to the minority subclass being exposed less in the optimization process and contributing less to the final model. This leads to a bias towards the majority subclass and suboptimal performance of the learned model. The difference between WCI and BCI lies in the lack of subclass labels. This deems common solutions for BCI that rely on sampling or cost-aware learning irrelevant for WCI as they rely on class labels.

\subsection{Challenge Construction}
\label{subsec:challenges}
To gain insights into how SSL and AL perform when the data challenges are present, we propose three datasets based on MNIST to reflect the challenges. We intentionally use MNIST as we can isolate any effects of the data challenges instead of the potential complexity of the learning task. 

\textbf{BCI-MNIST} 
We construct a between-class imbalanced version of MNIST (BCI-MNIST), where 50\% of the classes only contain approximately 10\% of the instances. \Cref{fig:bci_tsne} and \Cref{fig:bci_dist} illustrate the distribution of the imbalanced version in a 2D t-SNE plot, and a barplot respectively.

\textbf{BCS-MNIST} 
\Cref{fig:dirty_tsne} shows a 2D t-SNE-plot of an ambiguous version of MNIST proposed in ~\cite{mukhoti2021deterministic}. The dataset consists of normal MNIST and Ambiguous MNIST, containing a large fraction of ambiguous instances with questionable labels, thus increasing the class overlap. \Cref{fig:bcs_dist} shows the similarities of each instance to all instances not belonging to the same class. Compared to the original MNIST, the similarity among instances across classes is much higher. 
In our experiment, we select 5\% of instances from the original MNIST dataset and 95\% of instances from Ambiguous MNIST and refer to it as BCS-MNIST. 

\textbf{WCI-MNIST}
The WCI version of MNIST is constructed with the following procedure: 
(1) For each class, we create a sub-clustering using the K-means algorithm on the original input features with $k=300$. 
(2) For each constructed within-class cluster, we select one instance as the underrepresented subclass except for one majority subclass and remove the remaining instances.
(3) We copy all the instances within the majority subclass multiple times to restore the original training set size and randomly add Gaussian noise to create slightly different versions. 
The 2D t-SNE representation is shown in \Cref{fig:wci_tsne}. While the class boundaries are sharper than in \Cref{fig:bcs_tsne}, many subgroups within each class are spread around all the data space. \Cref{fig:wci_dist} shows the summed distance of each instance to the remaining instances of their respective class for MNIST (blue) and our constructed WCI-MNIST (orange). WCI-MNIST has more highly similar instances, and the number of medium distances is much smaller, resulting in a non-linear decrease in intra-class distances and higher within-class imbalance.

\section{Experiments}
\label{sec:experiments}
In this section, we evaluate established SSL methods combined with simple AL heuristics on the previously described challenges that ostensibly occur in real-world scenarios. We use the following experimental setup\footnote{See also \url{https://github.com/lmu-dbs/HOCOBIS-AL}}. \\
\textbf{Backbone and Training} For all experiments, we use a LeNet~\cite{lecun1989backpropagation} as backbone as is commonly used for digit recognition. We do not use a validation set as proposed in ~\cite{oliver2018realistic} since it is unrealistic to assume having a validation set when there is hardly any label information. Instead, we train the model for 50 epochs and use early stopping if the model reaches 99\% training accuracy following~\cite{badge}. The learning rate is set to $0.001$, and we do not use any scheduler. \\
\textbf{SSL} We include pseudo-labeling~\cite{lee2013pseudo} (PL) with a threshold of $0.95$ as baseline without consistency regularization. We further include Fixmatch~\cite{sohn2020fixmatch} as it is a well-established consistency regularization technique and Flexmatch~\cite{zhang2021flexmatch} as a strong method tackling confirmation bias~\cite{wang2022usb}. Furthermore, we report results on a plain supervised baseline (SPV).  \\
\textbf{Evaluation} 
We report average test accuracies over five random seeds for different labeling budgets. Initially, we select 20 labeled instances randomly. Then, we increase the labeled pool to budgets of 50, 100, 150, 200, and 250 labels. \\
\textbf{AL} We choose one representative from each of the described categories in \Cref{subsec:al} to better assess the strength and weaknesses of each acquisition type. 
We use margin uncertainty~\cite{settles2009active} as an uncertainty baseline. For representativeness, we perform k-means clustering on the latent features and select the instance closest to the centroid similar to \cite{prabhu2021active,gilhuber2022accelerating}.
As a coverage-based technique, we include the k-Center-Greedy method proposed in~\cite{sener2018active}.
For balanced sampling, we create a baseline that selects instances proportional to the sum of inverse class frequencies in the current labeled set and the corresponding prediction probability. Though this might not be a strong AL baseline in general, we expect to see a slight improvement in the BCI challenge. \\
\textbf{Datasets} We use the three constructed datasets explained in \Cref{subsec:challenges} to form the unlabeled pool, as well as the original MNIST. For testing, we use the original MNIST test set to ensure comparable results.

\begin{figure}[ht]
    \centering
    \begin{subfigure}[b]{0.32\textwidth}
        \includegraphics[width=\textwidth]{img/tsne_BCI.png}
        \caption{t-SNE BCI-MNIST.}
        \label{fig:inter_tsne}
    \end{subfigure}
    \hfill
    \begin{subfigure}[b]{0.32\textwidth}
        \includegraphics[width=\textwidth]{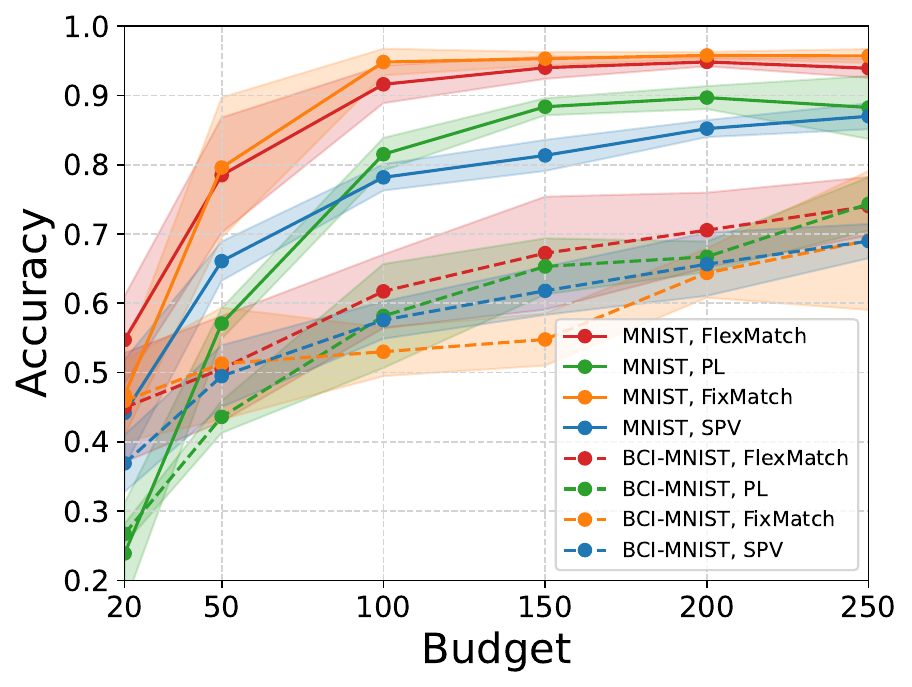}
        \caption{Random.}
        \label{fig:inter_curves}
    \end{subfigure}
    \begin{subfigure}[b]{0.32\textwidth}
        \includegraphics[width=\textwidth]{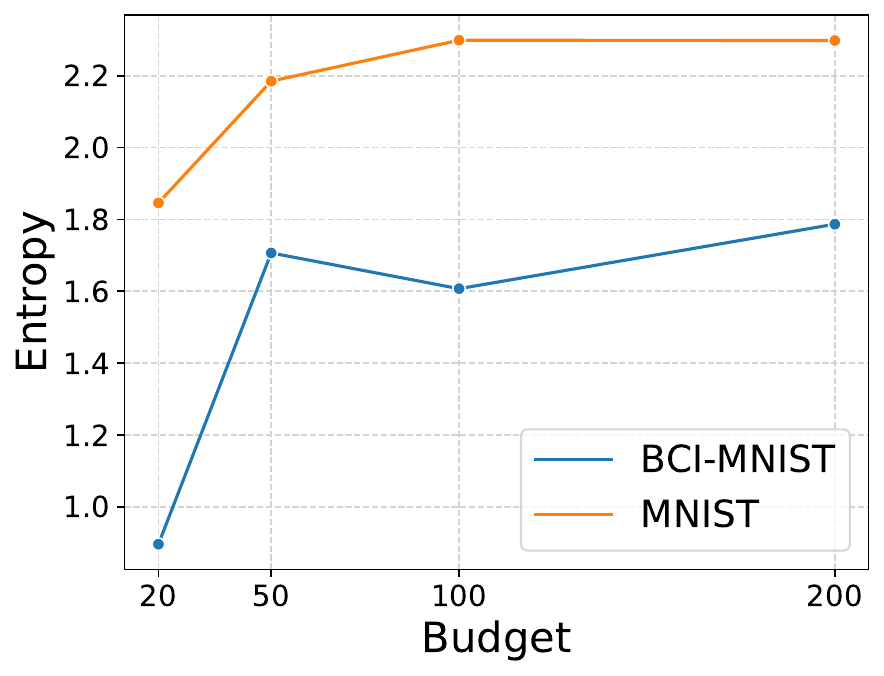}        
        \caption{Entropy over PL.}
        \label{fig:pl_inter}
    \end{subfigure}
    \hfill
    \begin{subfigure}[b]{0.32\textwidth}
        \includegraphics[width=\textwidth]{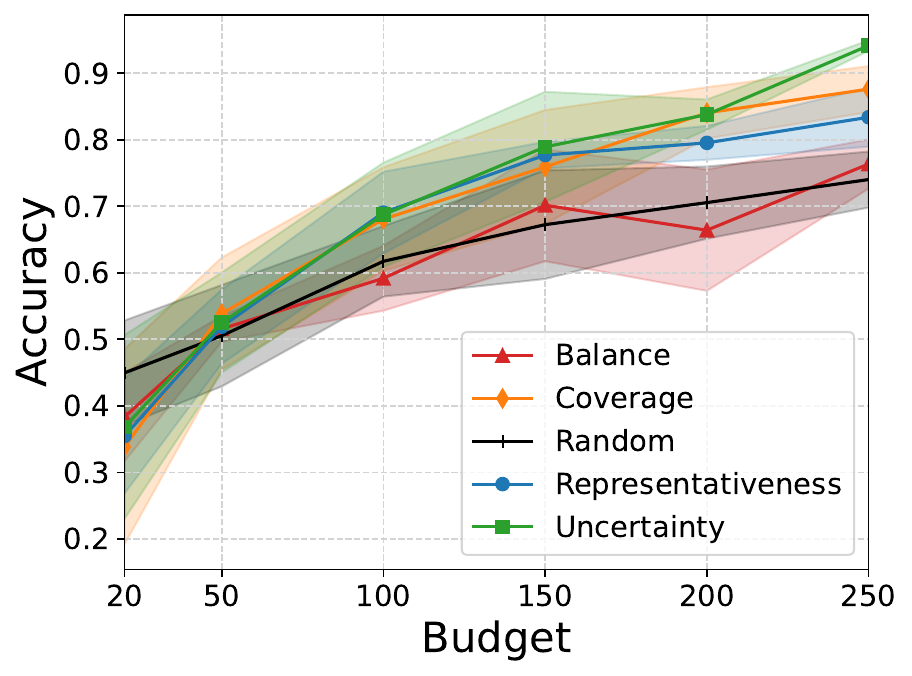}
        \caption{AL Flexmatch.}
        \label{fig:al_flexmatch_inter}
    \end{subfigure}
    \hfill  
    \begin{subfigure}[b]{0.32\textwidth}
        \includegraphics[width=\textwidth]{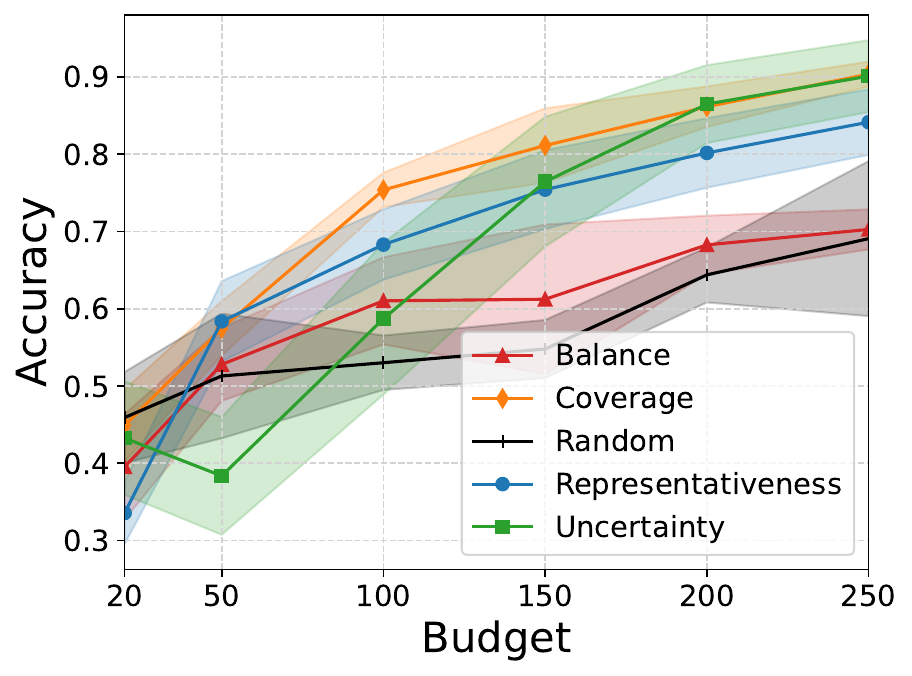}
        \caption{AL Fixmatch.}
        \label{fig:al_fixmatch_inter}
    \end{subfigure}
    \hfill 
    \begin{subfigure}[b]{0.32\textwidth}
        \includegraphics[width=\textwidth]{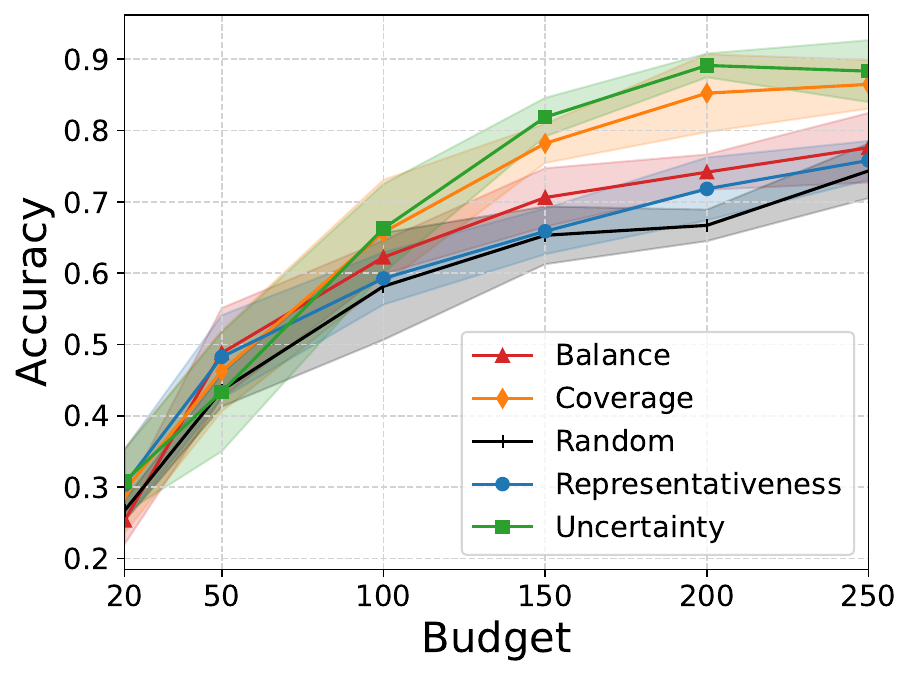}
        \caption{AL PL.}
        \label{fig:al_pl_inter}
    \end{subfigure}
    \caption{(a) t-SNE of BCI-MNIST challenge. (b) Average test accuracy of all learners evaluated on BCI-MNIST (dashed) and MNIST (solid). In (d), we observe for BCI-MNIST, the entropy over the selected pseudo-labels falling over the threshold for each class is much smaller. This indicates that the distribution of selected pseudo-labels for BCI-MNIST is more imbalanced, repeatedly confirming the imbalance. (d), (e) and (f) show the selected AL curves for Flexmatch, Fixmatch, and PL compared to random sampling (black).}
    \label{fig:challenges}
\end{figure}
\subsection{Experiment "BCI-MNIST"}
\Cref{fig:inter_curves} depicts the average accuracy of supervised learning (SPV, blue), pseudo-labeling (PL, green),  Fixmatch (orange), and Flexmatch (red) for different labeling budgets on MNIST (solid) and BCI-MNIST (dashed) with random labeling. BCI has a severe impact on the performance of all learners. However, Fixmatch is affected most and even performs worse than SPV. Since training takes much longer for SSL, \cite{oliver2018realistic} argue that these methods should clearly outperform SPV to be considered useful. This is no longer true in our experiment, even on a simple task like MNIST. \Cref{fig:pl_inter} visualizes the entropy over the number of pseudo-labeled instances per class that Fixmatch would choose for training for BCI-MNIST (blue) and MNIST (orange). 
On MNIST the entropy is much higher, indicating that the distribution over the classes is more uniformly distributed. The problem is not only that the selected labeled data is imbalanced, but the chosen pseudo-labels repeatedly \emph{confirm} the imbalance, such that the underrepresented classes get even more underrepresented. 

However, the AL curves in \Cref{fig:al_flexmatch_inter,fig:al_fixmatch_inter,fig:al_pl_inter} demonstrate that the choice of data selection methods has a substantial impact on the performance of each learner. Fixmatch largely benefits from coverage-based sampling, representative sampling, and uncertainty sampling for later iterations. For the final budget of 250, the gap between coverage and uncertainty acquisition and random selection is around 20\%. PL and Flexmatch also greatly benefit from coverage and uncertainty sampling. Coverage sampling is even able to restore the accuracy achieved on MNIST with random sampling, yielding 88.3\% for PL and 94.1\% for Flexmatch. Interestingly, balanced sampling is not among the best active methods. 
Even though the performance is slightly better than random sampling, the other methods are much stronger. 
This is probably because balanced sampling without the combination of any other method does select less informative and more redundant information. 

\begin{figure}[ht]
    \centering
    \begin{subfigure}[b]{0.32\textwidth}
        \includegraphics[width=\textwidth]{img/tsne_BCS.png}
        \caption{t-SNE BCS.}
        \label{fig:dirty_tsne}
    \end{subfigure}
    \hfill 
    \begin{subfigure}[b]{0.32\textwidth}
        \includegraphics[width=\textwidth]{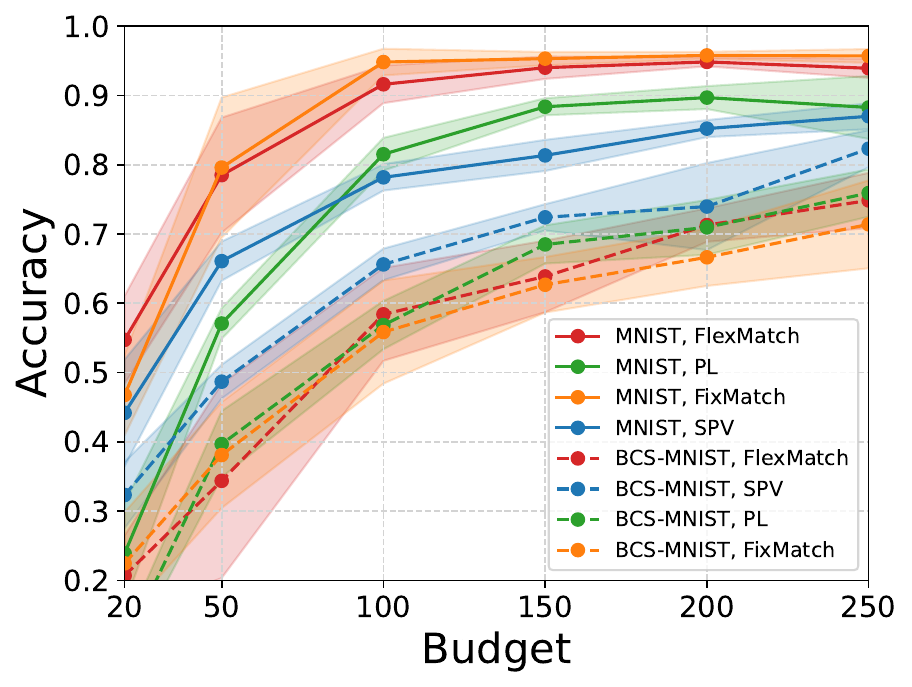}
        \caption{Random.}
        \label{fig:dirty_curve}
    \end{subfigure}
    \begin{subfigure}[b]{0.32\textwidth}
        \includegraphics[width=\textwidth]{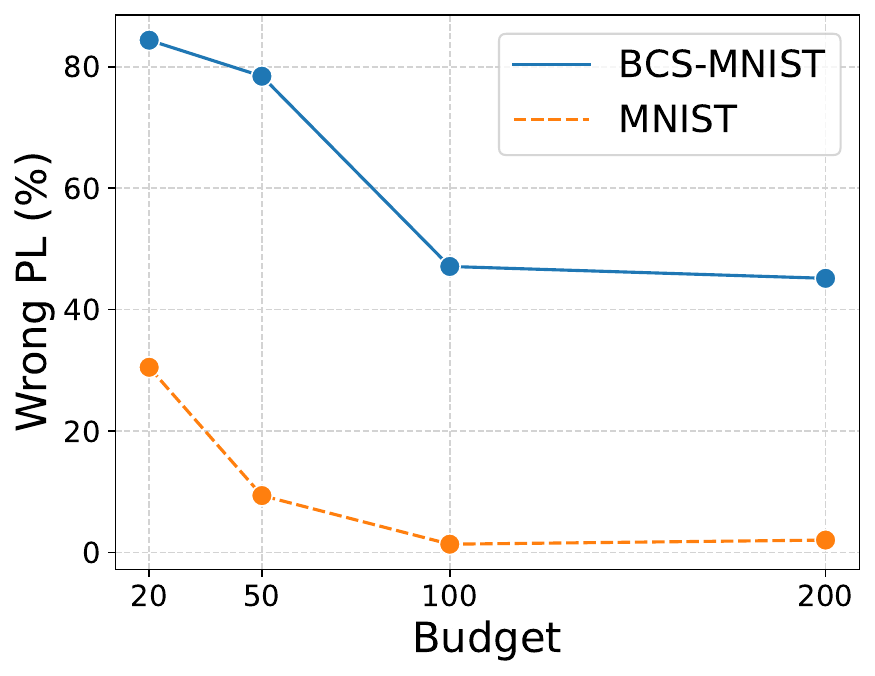}
        \caption{Incorrect PL.}
        \label{fig:pl_dirty}
    \end{subfigure}
    \hfill 
    \begin{subfigure}[b]{0.32\textwidth}
        \includegraphics[width=\textwidth]{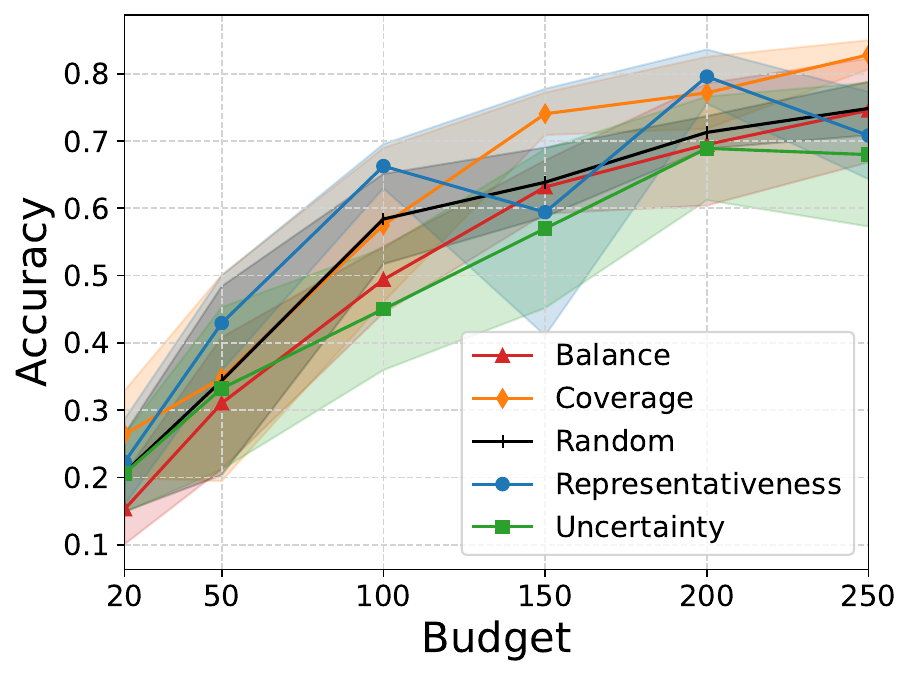}
        \caption{AL Flexmatch.}
        \label{fig:al_flexmatch_dirty}
    \end{subfigure}
    \hfill 
    \begin{subfigure}[b]{0.32\textwidth}
        \includegraphics[width=\textwidth]{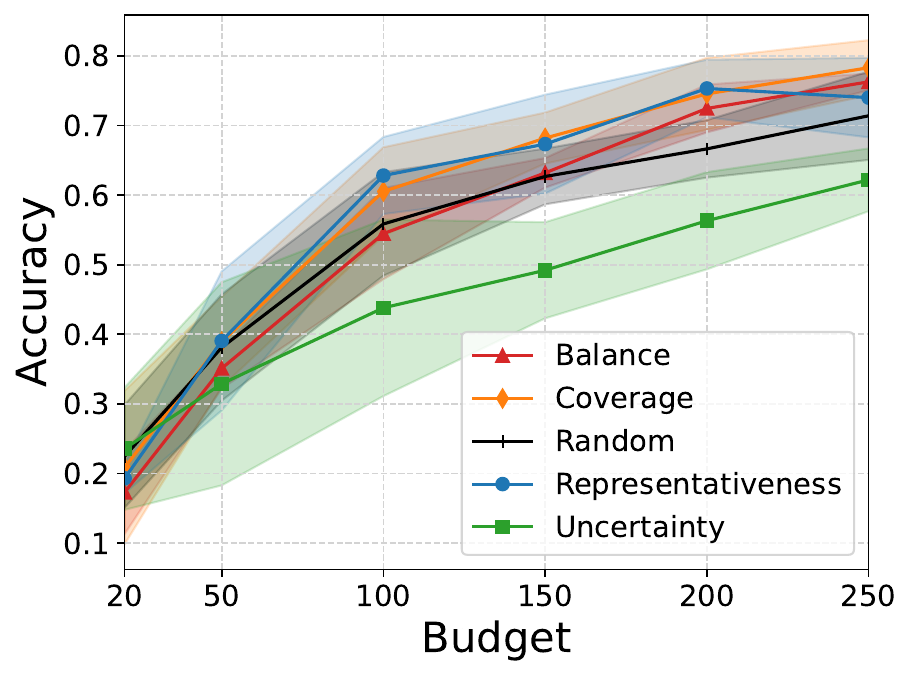}
        \caption{AL Fixmatch.}
        \label{fig:al_fixmatch_dirty}
    \end{subfigure}
    \hfill
    \begin{subfigure}[b]{0.32\textwidth}
        \includegraphics[width=\textwidth]{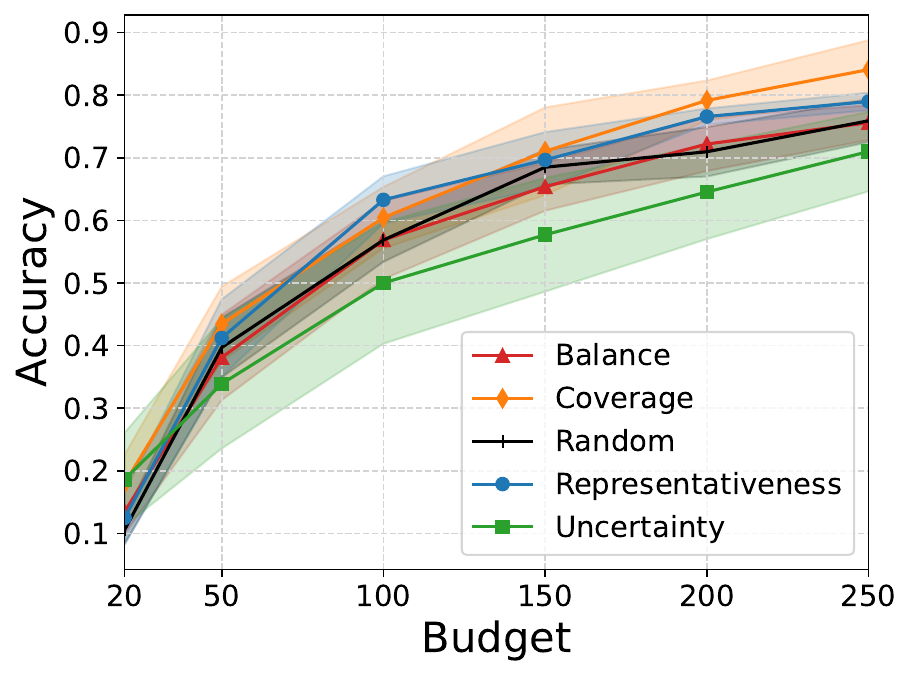}
        \caption{AL PL.}
        \label{fig:al_pl_dirty}
    \end{subfigure}
    
    \caption{(b) Average test accuracy of all learners with random selection for BCS-MNIST (dashed) and MNIST (solid). (c) shows the amount of wrongly predicted pseudo-labels falling over the threshold using Fixmatch for BCS-MNIST is much larger than for MNIST. (d), (e) and (f) show the AL curves for Flexmatch, Fixmatch, and PL compared to random sampling (black).}
    \label{fig:BCS_all}
\end{figure}

\subsection{Experiment "BCS-MNIST"}
\Cref{fig:dirty_curve} illustrates the learning curves for the learners on MNIST and BCS-MNIST. All methods suffer, but Fixmatch clearly suffers the most and is no longer better than plain supervision. In this scenario, there is no additional benefit of exploiting the unlabeled pool, but the training times are multiple times larger. \Cref{fig:pl_dirty} illustrates the fraction of wrong pseudo-labels surpassing the threshold when training Fixmatch on MNIST (orange) and BCS-MNIST (blue). Over 40\% of the predicted pseudo-labels over the threshold are wrong up to a labeling budget of 200 instances. 
\Cref{fig:al_flexmatch_dirty,fig:al_fixmatch_dirty,fig:al_pl_dirty} denote the learning curves of Flexmatch, Fixmatch, and PL when increasing the labeled pool actively. Notably, all learners benefit from coverage-based sampling. Representative sampling is beneficial for Fixmatch. This method promotes instances representative of a certain class or region and probably selects instances that are less ambiguous for training. However, as expected, employing the uncertainty baseline in this context proves to be a poor choice. 
The strategy lacks the ability to differentiate between aleatoric and epistemic uncertainty, leading to the selection of many ambiguous instances, further misleading the training.

\begin{figure}[ht]
    \centering
    \begin{subfigure}[b]{0.32\textwidth}
        \includegraphics[width=\textwidth]{img/tsne_WCI.png}
        \caption{t-SNE WCI.}
        \label{fig:intra_tsne}
    \end{subfigure}
    \hfill
    \begin{subfigure}[b]{0.32\textwidth}
        \includegraphics[width=\textwidth]{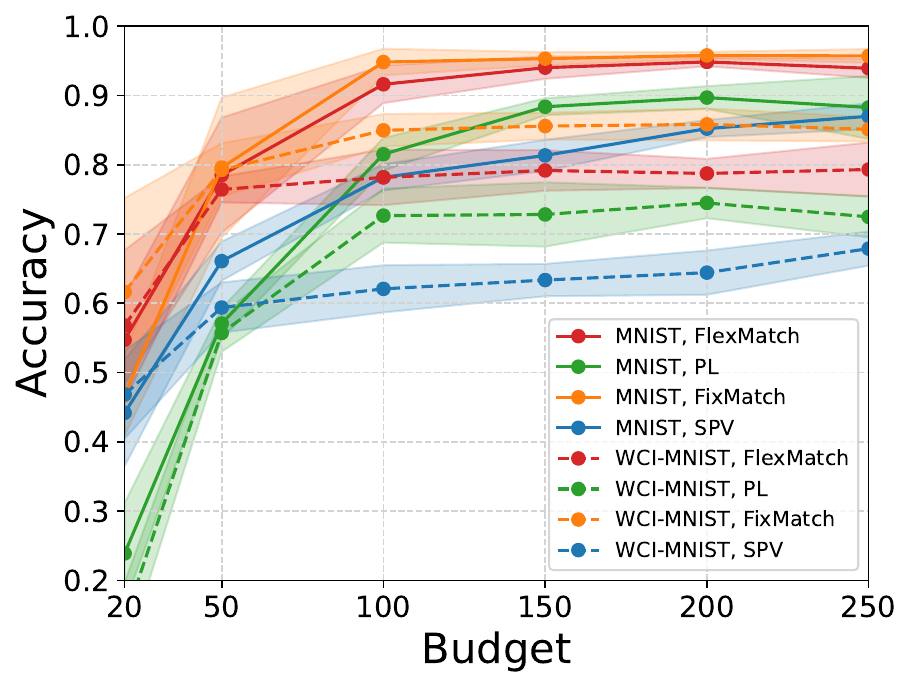}
        \caption{Random.}
        \label{fig:intra_random}
    \end{subfigure}
    \hfill
    \begin{subfigure}[b]{0.32\textwidth}
        \includegraphics[width=\textwidth]{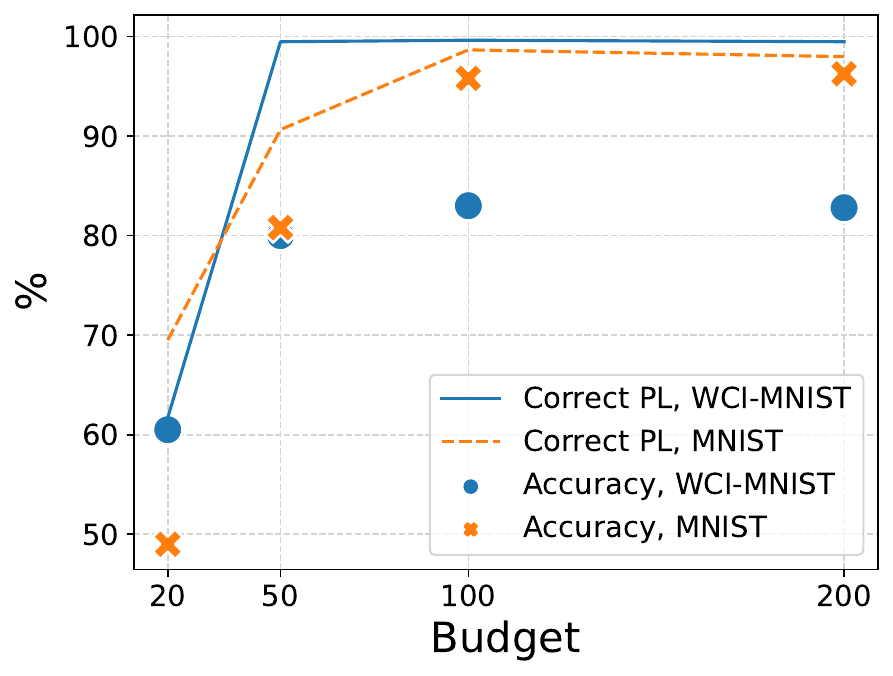}
        \caption{PL vs. Accuracy.}
        \label{fig:intra_confirmationbias}
    \end{subfigure}
    
    \begin{subfigure}[b]{0.32\textwidth}
        \includegraphics[width=\textwidth]{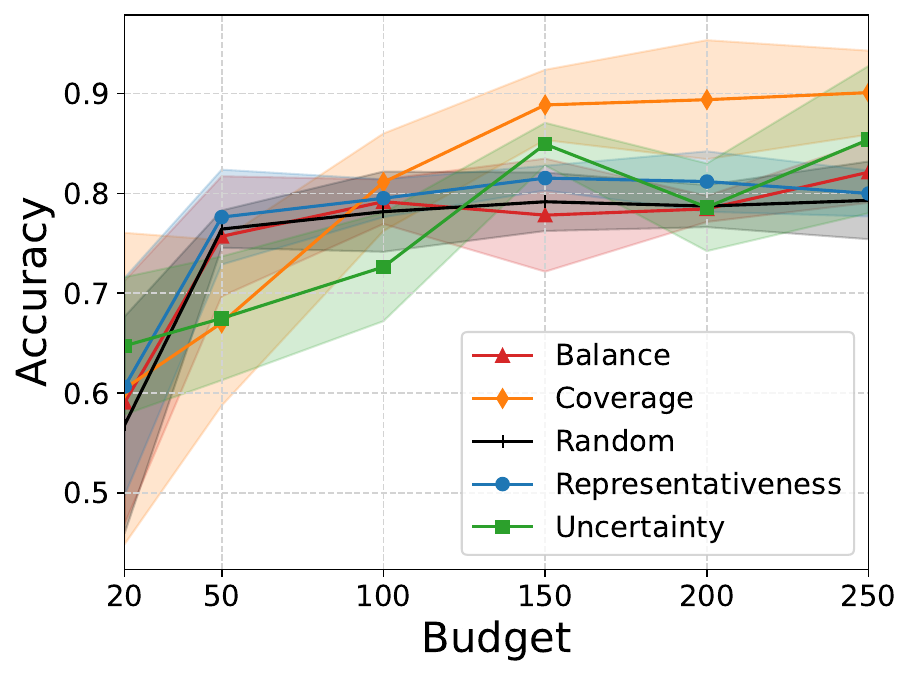}
        \caption{AL Flexmatch.}
        \label{fig:intra_flexmatch}
    \end{subfigure}
    \hfill  
    \begin{subfigure}[b]{0.32\textwidth}
        \includegraphics[width=\textwidth]{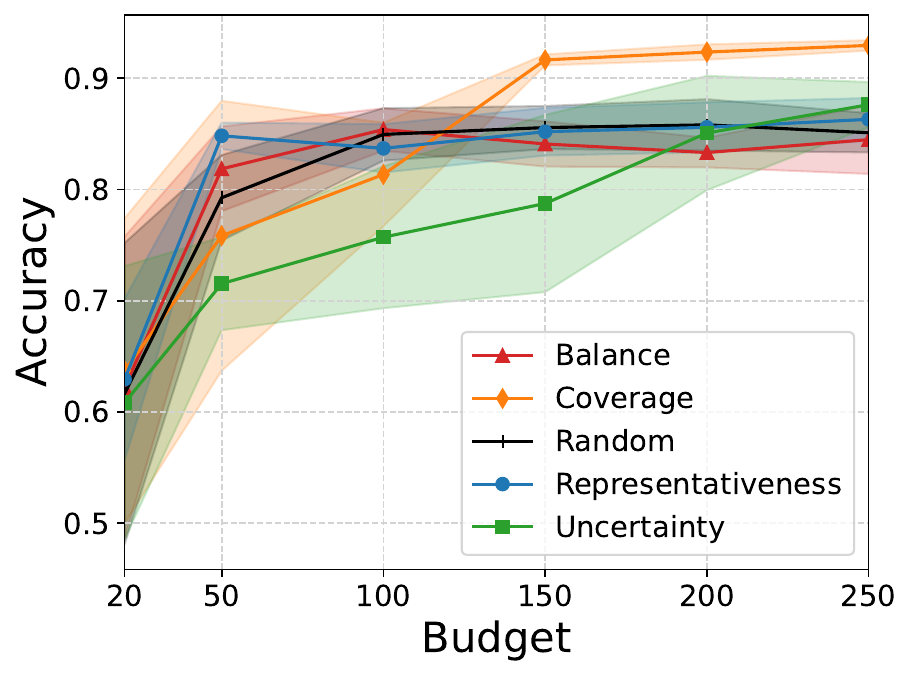}
        \caption{AL Fixmatch.}
        \label{fig:al_intra_fixmatch}
    \end{subfigure}
    \hfill
    \begin{subfigure}[b]{0.32\textwidth}
        \includegraphics[width=\textwidth]{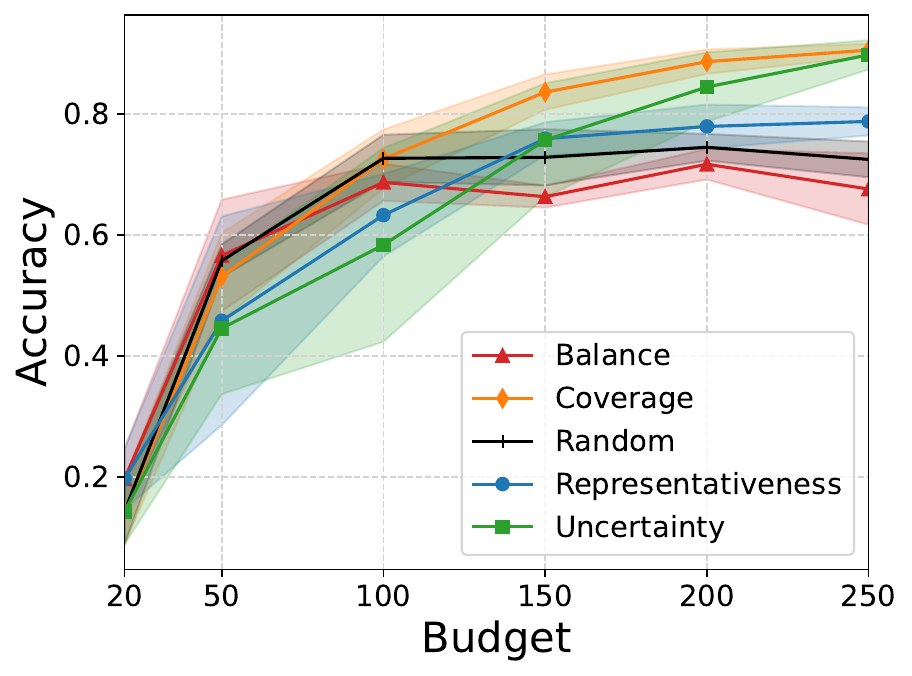}
        \caption{AL PL.}
        \label{fig:al_intra}
    \end{subfigure}
    
    \caption{(b) Average test accuracy of all learners for WCI-MNIST (dashed) and MNIST (solid). In (d), we observe that even though more pseudo-labels are chosen correctly using Fixmatch for WCI-MNIST (blue line), the test accuracy is much smaller (blue markers) than for MNIST (orange) because only the same concepts are confirmed over and over again. (d), (e) and (f) show the selected AL curves for Flexmatch, Fixmatch, and PL compared to random sampling (black).}
    \label{fig:intra_challenges}
\end{figure}

\subsection{Experiment "WCI-MNIST"}
\Cref{fig:intra_random} shows that for WCI-MNIST, the accuracy of all learners stagnates around 10\% to 15\% earlier compared to MNIST. Using random sampling does not find the underrepresented diverse instances, and only the same concepts are entrenched and further confirmed over the training procedure. 
Even though the correctness ratio of the pseudo-labels surpassing the threshold using Fixmatch is larger for WCI-MNIST than for MNIST, the achieved mean test accuracy stops at roughly 82\% (see \Cref{fig:intra_confirmationbias}). 

However, using AL, we can find more diverse and valuable instances than the already known concepts and reach a better final accuracy overall for SSL (see \Cref{fig:al_intra_fixmatch,fig:al_intra,fig:intra_flexmatch}). Especially coverage-based sampling seems to be a viable choice. For PL, the final average accuracy using uncertainty-based and coverage-based sampling on WCI-MNIST is even equally good as the performance on the original MNIST using random sampling. In the early stages, uncertainty sampling is the worst method probably because it lacks diversity aspects, and the predictions in early iterations might not be very reliable. However, for the final budget, uncertainty sampling matches or surpasses most other methods.  The representative baseline focuses on instances that are most central in clusters, probably resulting in only selecting the already known and easy-to-classify concepts lacking novel information and does not outperform random sampling in most situations.

\begin{table}[ht!]
\centering
\caption{Average test accuracy for SPV, Fixmatch, PL, and Flexmatch for BCI-MNIST, BCS-MNIST, and WCI-MNIST for all sampling methods and budgets 50 and 250 (L). \textbf{Bold} and \textcolor{red}{red} numbers indicate column-wise best- and worst-performing methods, respectively.}
\resizebox{\textwidth}{!}{\begin{tabular}{lcccccc|cccccc|cccccc|cccccc}
\toprule
     & \multicolumn{6}{c}{Supervised} & \multicolumn{6}{c}{Fixmatch} & \multicolumn{6}{c}{Pseudo-Labeling} & \multicolumn{6}{c}{Flexmatch} \\

\midrule   
     &   \multicolumn{2}{c}{BCI}    & \multicolumn{2}{c}{BCS}    & \multicolumn{2}{c}{WCI}       & \multicolumn{2}{c}{BCI}   & \multicolumn{2}{c}{BCS}   & \multicolumn{2}{c}{WCI}        & \multicolumn{2}{c}{BCI}     & \multicolumn{2}{c}{BCS}     & \multicolumn{2}{c}{WCI}   & \multicolumn{2}{c}{BCI}     & \multicolumn{2}{c}{BCS}     & \multicolumn{2}{c}{WCI}  \\
\midrule 
 L  & 50 & 250 & 50 & 250 & 50 & 250 & 50 & 250 & 50 & 250 & 50 & 250 & 50 & 250 & 50 & 250 & 50 & 250 & 50 & 250 & 50 & 250 & 50 & 250 \\
\midrule
Rnd  & 49.4 & 68.9  & 48.6 & 82.3  & 59.3 & 67.8  & 51.2 & 69.0  & 38.1 & 71.3  & 79.2 & 85.1  & 43.6 & 74.3  & 39.6 & 75.9  & 55.7 & 72.4  & \textcolor{red}{50.5} & \textcolor{red}{74.0}  & 34.3 & 74.8  & 76.4 & \textcolor{red}{79.3} \\
Unc  & 52.0 & \textbf{82.5}  & 53.0 & 80.5  & \textcolor{red}{47.8} & 70.0  & \textcolor{red}{38.3} & 90.1  & \textcolor{red}{32.8} & \textcolor{red}{62.1}  & \textcolor{red}{71.5} & 87.6  & \textcolor{red}{43.3} & \textbf{88.3}  & \textcolor{red}{33.8} & \textcolor{red}{70.9}  & \textcolor{red}{44.5} & 89.7  & 52.5 & \textbf{94.1}  & 33.2 & \textcolor{red}{67.9}  & 67.4 & 85.3 \\
Cov  & \textcolor{red}{47.9} & 79.2  & \textbf{55.2} & \textbf{83.8}  & \textbf{63.0} & \textbf{87.6}  & 57.4 & \textbf{90.3}  & 38.9 & \textbf{78.2}  & 75.8 & \textbf{92.9}  & 46.2 & 86.4  & \textbf{43.4} & \textbf{84.0}  & 53.0 & \textbf{90.5}  & \textbf{53.8} & 87.6  & 34.7 & \textbf{82.8}  & \textcolor{red}{67.0} & \textbf{90.1} \\
Bal  & 48.2 & 68.7  & 50.6 & 78.5  & 58.0 & \textcolor{red}{64.1}  & 52.7 & 70.2  & 35.0 & 76.2  & 81.8 & \textcolor{red}{84.4}  & \textbf{48.8} & 77.6  & 38.0 & 75.5  & 56.5 & \textcolor{red}{67.5}  & 51.5 & 76.3  & \textcolor{red}{31.0} & 74.5  & 75.7 & 82.1 \\
Rep  & \textbf{54.7} & \textcolor{red}{66.8}  & \textcolor{red}{47.7} & \textcolor{red}{75.8}  & 61.6 & 66.1  & \textbf{58.3} & 84.1  & 39.0 & 73.9  & \textbf{84.8} & 86.3  & 48.2 & 75.8  & 41.2 & 78.9  & 45.7 & 78.7  & 51.8 & 83.3  & \textbf{42.9} & 70.8  & \textbf{77.6} & 79.9 \\

\bottomrule
\end{tabular}}
\label{table:summary}
\end{table}

\section{Key Findings}
\label{sec:discussion}
\Cref{table:summary} shows the average test accuracies of SPV, Fixmatch, PL, and Flexmatch on BCI-MNIST, BCS-MNIST, and WCI-MNIST for all AL heuristics compared to random sampling, where bold and red numbers indicate best- and worst-performing methods per column respectively for 50 and 250 labeled instances. Our key findings can be summarized as follows:

\begin{itemize}
    \item For all introduced data challenges, the SSL methods suffer from confirmation bias. There is no consistent winner among all query strategies, but random sampling is never the best query method for the SSL methods when faced with BCS, WCI, and BCI. This provides empirical evidence that AL is a useful tool to overcome confirmation bias in SSL.
    % \item The worst and strongest methods in the supervised setting do not necessarily translate to the strongest and worst methods for SSL.  % TODO: Re-add?
    \item In the early stages, representative sampling is often beneficial. In contrast, uncertainty sampling usually performs better in later iterations where model predictions are more reliable. As expected, uncertainty sampling is not a good choice for BCS since it queries from overlapping, confusing regions.  
    \item Coverage sampling is often the best strategy for SSL methods. We assume that is because more diverse queried instances bring in new aspects to the data, and the easier concepts can already be learned by pseudo-labeling and consistency regularization.
    \item Our balance baseline often performs on par with random selection. However, for the BCI challenge, it yields slightly better results. We conclude that it should mainly be used in combination with other selection heuristics. 
    \item Overall, the most challenging dataset for SSL and AL is BCS-MNIST. By using AL, we can mitigate confirmation bias more effectively for the challenges BCI and WCI compared to random sampling. 
\end{itemize}

\section{Conclusion}
\label{sec:conclusion}
In this work, we study the real-world transferability of critique points on the combination of SSL and AL on benchmark datasets.
Our experiments show that AL is a useful tool to overcome confirmation bias in various real-world challenges. However, it is not trivial to determine which AL method is most suitable in a real-world scenario. This study is limited to providing insights into confirmation bias in SSL when confronted with between-class imbalance, between-class similarity, and within-class similarity and the potential of simple AL heuristics. 
In the future, we intend to extend our experiments to a broader range of datasets, with a strong focus on real-world examples.
Moreover, we aim to include existing hybrid AL methods in our evaluation and to design a robust active semi-supervised method capable of consistently overcoming confirmation bias in SSL on diverse challenges.

\section*{Acknowledgements}
This work was supported by the Bavarian Ministry of Economic Affairs, Regional Development and Energy through the Center for Analytics – Data – Applications (ADA-Center) within the framework of BAYERN DIGITAL II (20-3410-2-9-8) as well as the German Federal Ministry of Education and Research (BMBF) under Grant No. 01IS18036A
%
% ---- Bibliography ----
%
% BibTeX users should specify bibliography style 'splncs04'.
% References will then be sorted and formatted in the correct style.
\bibliographystyle{splncs04}
\bibliography{main}
\clearpage

\appendix

\counterwithin{figure}{section}
\counterwithin{table}{section}

\section{Literature Overview}
\begin{table}[!h]
\begin{tabular}{lp{6cm}cccc}
\toprule
\textbf{Paper} & \textbf{Real-World Considerations} & \textbf{BCI} & \textbf{WCI} & \textbf{BCS} & \textbf{SSL}\\  \toprule
Kothawade et al. \cite{Kothawade2021SIMILARSI} & Imbalance or rare classes, out-of-distribution data, redundancy in the unlabeled set & \checkmark & \checkmark & \\
Park et al. \cite{Park2022MetaQueryNetRP}     & Open-set noise \\
Elenter et al. \cite{Elenter2022ALD}  & Dataset redundancy in STL-10 & & \checkmark & \\
Kirsch et al. \cite{Kirsch2019BatchBALDEA}    & Repetition in MNIST & & \checkmark & \\
Liang et al. \cite{Liang2020ALICEAL}    & Incorporation with natural language explanation & & & \checkmark             \\
Hacohen et al. \cite{Hacohen2022ActiveLO}  & An imbalanced subset of CIFAR-10 & \checkmark & & & \checkmark\\
Zhang et al. \cite{Zhang2022GALAXYGA}    & Extreme class imbalance & \checkmark & &                 \\
Beluch et al. \cite{Beluch2018ThePO}    &  Highly class-imbalanced diabetic retinopathy dataset (in medical diagnosis) & \checkmark & &                 \\
Ning et al. \cite{Ning2022ActiveLF}     & Open-set annotation problem & & & \\
Munjal et al. \cite{Munjal2020TowardsRA}    & Class imbalance & \checkmark & & \\
Zhang et al. \cite{Zhang2022BoostMISBM}     & Poor data utilization and missing informative sample in medical data & & & & \checkmark \\
Choi et al. \cite{Choi2020VaBALIC}     & The heavily imbalanced NEU dataset & \checkmark & & \\
Zhang et al. \cite{Zhang2020StateRelabelingAA}    & Class imbalance in Caltech-101 & \checkmark & \\
Gudovskiy et al. \cite{Gudovskiy2020DeepAL} & Biased class imbalance & \checkmark & & & \checkmark\\
Wang et al. \cite{wang2022boosting}      & Imbalanced data & \checkmark &                                    &                 \\
Ning et al. \cite{Ning2021ImprovingMR}     & Unexpected noise  &     &                                    & \checkmark \\
Sinha et al. \cite{Sinha2019VariationalAA}     & Noisy data caused by an inaccurate oracle  &     &                                    & \checkmark & \checkmark\\
Mullapudi et al. \cite{Mullapudi2021LearningRC} & Imbalanced data & \checkmark & \checkmark & & \checkmark\\
Du et al. \cite{Du2021ContrastiveCF}       & Class distribution mismatch & \\
Yi et al. \cite{Yi2022UsingSP}        & Imbalanced data, cold-start problem & \checkmark & & \\
Kim \& Shin \cite{Kim2022InDO}     & Redundancy and highly similar samples  &     & \checkmark &                 \\
Shao et al. \cite{Shao2019LearningTS}     & Highly imbalanced classes and cold-start problem & \checkmark &      &                 \\
Zhang et al. \cite{Zhang2018SimilarityBasedAL}    & Class imbalance & \checkmark & & \\ \bottomrule
\end{tabular}
\caption{Overview of studies evaluated in realistic scenarios. BCI is between-class imbalance, WCI is within-class imbalance and BCS is between-class similarity. SSL denotes if a study combines AL with SSL.}
\label{tab:study-overview}
\end{table}

% \begin{figure}[!h]
%     \centering
%     \includegraphics[width=0.4\textwidth]{img/Appendix_benchmark_dataset.pdf}
%     \caption{The number of papers experimenting on the well-established datasets}
%     \label{fig:my_label}
% \end{figure}  

\end{document}